\documentclass[symmetry,article,submit,moreauthors,pdftex]{Definitions/mdpi} 
\usepackage{subfigure}
\usepackage{amssymb}
\usepackage{makecell}
\usepackage{url}

\firstpage{1} 
\pubvolume{xx}
\issuenum{1}
\articlenumber{5}
\pubyear{2021}
\copyrightyear{2021}
\history{Received: date; Accepted: date; Published: date}

\Title{PointSCNet: Point Cloud Structure and Correlation Learning Based on Space Filling Curve Guided Sampling}

\Author{Xingye Chen $^{1}$, Yiqi Wu $^{1,2*}$, Wenjie Xu $^{1}$ , Jin Li $^{1}$, Huaiyi Dong $^{1}$, and Yilin Chen $^{2,3}$}

\AuthorNames{Firstname Lastname, Firstname Lastname, Firstname Lastname, Firstname Lastname and Firstname Lastname}

\address{%
$^{1}$ \quad School of Computer Science, China University of Geosciences, Wuhan 430074, China\\
$^{2}$ \quad Hubei Key Laboratory of Intelligent Robot (Wuhan Institute of Technology), Wuhan 430205, China\\
$^{3}$ \quad School of Computer Science and Engineering, Wuhan Institute of Technology, Wuhan 430205, China\\
}

\corres{Correspondence: wuyq@cug.edu.cn}

\abstract{The geometrical structure and internal local region relationship, such as symmetry, regular array, junction and so on, are essential for understanding a 3D shape. This paper proposes a point cloud feature extraction network named PointSCNet, to capture the geometrical structure information and local region correlation information of the point cloud. The PointSCNet consists of three main modules: the space filling curve guided sampling module, the information fusion module and the channel-spatial attention module. The space filling curve guided sampling module uses Z-order curve coding to sample points which contain geometrical correlation. The information fusion module uses a correlation tensor and a set of skip connections to fuse the structure and correlation information. The channel-spatial attention module enhances the representation of key points and crucial feature channels to refine the network. The proposed PointSCNet is evaluated on shape classification and part segmentation tasks. The experiment results demonstrate that the PointSCNet outperforms or is on par with state-of-the-art methods by learning structure and correlation of point cloud effectively. The source code of the PointSCNet is available at \url{https://github.com/Chenguoz/PointSCNet}.}

\begin{document}

\section{Introduction}
\begin{figure}[h]
	\begin{center}
		\includegraphics[width=0.7\textwidth]{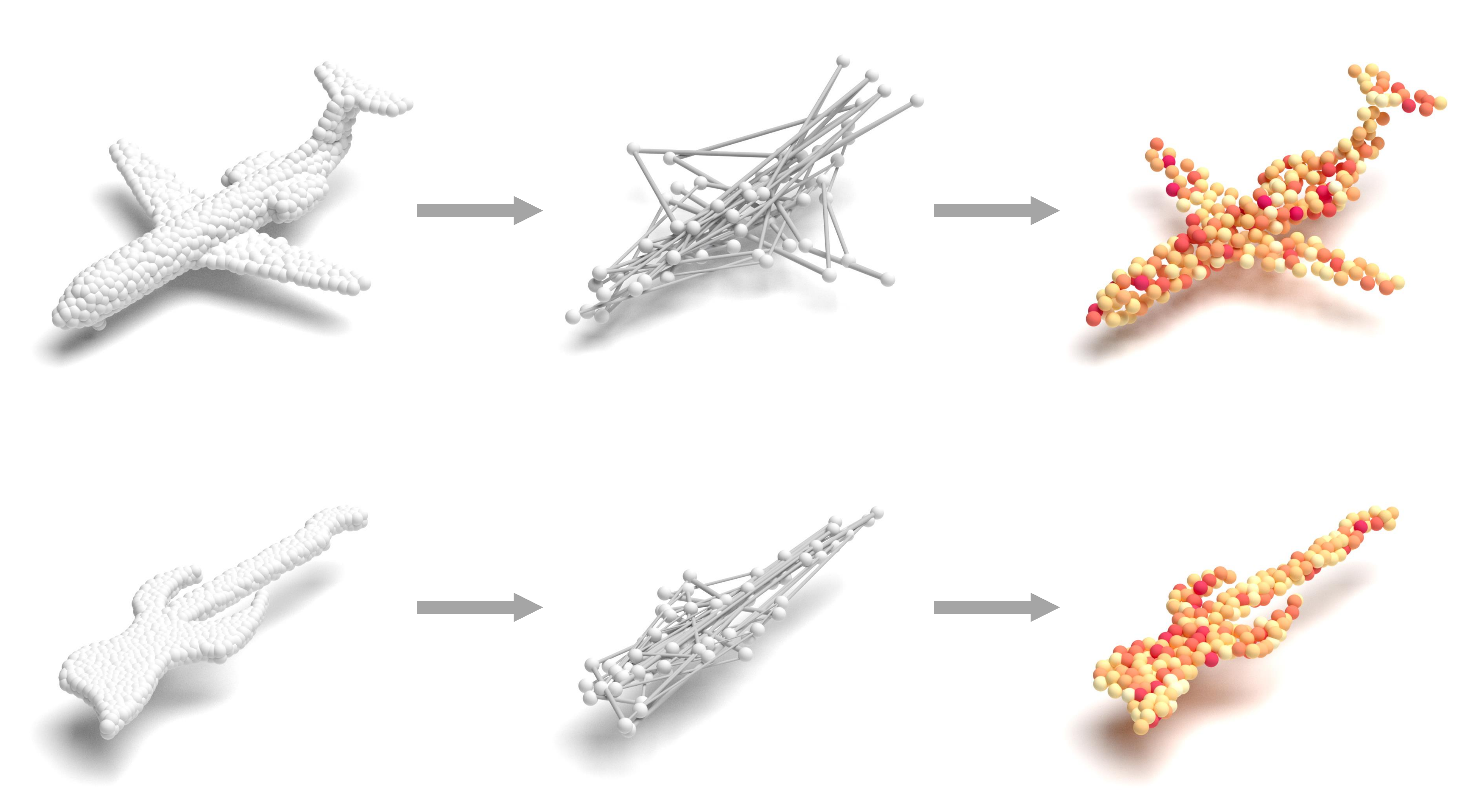}
	\end{center}
	\caption{Learning structure and correlation on point cloud based on space filling curve guided sampling. Columns shown from left to right are the original input point cloud, points sampled by the Z-order space filling curve, and the point cloud heat map based on response of points to the proposed network, respectively.} 
	\label{fig:point-order-attention}
\end{figure}
Point cloud is an ubiquitous form of 3D shape and suitable for countless applications in computer graphics due to its accessibility and expressiveness for 3D representation. 
The points are captured from the surface of objects by equipments such as 3D scanner, Light Detection and Ranging (LiDAR) or RGB-D camera, or sampling from other 3D representations~\cite{guo2013rotational}. 
While containing rich information about the surface, structure and shape of 3D objects, they are unlikely to be ordered and structured data as images which are arranged on regular pixel grids.
Hence, although many classical deep neural networks have shown tremendous success in image processing, there are still a lot of challenges when it comes to deep learning methods for point cloud~\cite{guo2020deep}.

To coordinate these incompatibilities, an intuitive idea is transforming the point cloud into a structured representation. 
Earlier multi-view methods try to project the 3D object into multiple view-wise images to fit 2D image processing approaches~\cite{qi2016volumetric,su2015multi,chen2017multi,yu2018multi,yang2019learning}.
On the other hand, volumetric methods voxelize the point cloud to a regular 3D grid representation and adopt extensions of the 2D networks, such as 3D Convolutional Neural Network (CNN), for feature extraction~\cite{maturana2015voxnet}. 
Moreover, some following voxel-based researches introduce certain data structure(such as octree) to reorganize the input shape~\cite{riegler2017octnet,wang2017cnn,le2018pointgrid}.  
While achieving impressive performances at the time, these methods are often considered to have some inevitable shortcomings, such as losing 3D geometric information during the 2D projection or high computational and memory costs when processing voxels. 

Against this backdrop, nowadays researches on directly consuming raw point clouds via end to end networks become increasingly popular.
The well-known PointNet~\cite{qi2017pointnet} and subsequent PointNet++~\cite{qi2017pointnet++} are the pioneer works of direct point cloud processing based on deep learning methods. 
The introduce of symmetric function reflected by the networks adapts to the inherent characteristics of the 3D points set.
Inspired by PointNet and Pointnet++, many following researches adopt the idea for points feature extraction or encoding to achieve the permutation invariance of point clouds~\cite{duan2019structural,yin2018p2p,yang2019modeling,sarode2019pcrnet}.

The basic methodology of these point-based network is exacting point-wise high dimensional information and then aggregating a local or global representation of the point cloud for downstream tasks. 
Followed-up researches based on this idea have demonstrate that the hierarchical structure with the subset abstraction procedure is effective for point cloud reasoning. 
It has been figured out that sampling central subsets of input points is essential for the hierarchical structures~\cite{lin2017structured,thabet2019mortonnet}.
However, the most popular sampling and grouping method, Farthest Point Sampling (FPS) and K-Nearest Neighbor (KNN), is based on low-dimension Euclidean distance exclusively, without sufficient consideration to the semantically high-level correlations of the points and their surrounding neighbors.

In the real world, there are inherent correlation between local regions of 3D objects, especially for those Computer Aided Design (CAD) models or industrially manufactured products~\cite{wu2020grid}, such as the symmetric wings design of an airplane, the regular arrays of wheels for a car or the distinct structure between the collar, sleeves and body part of a shirt. These geometric correlations of local regions  play crucial role in 3D object understanding and are significant for typical point cloud processing tasks such as shape classification and part segmentation.      

Besides, in the procedure of high dimensional information extraction, a basic and effective approach is using shared Multi-Layer Perceptron (MLP) or 1D CNN to project the input feature to a high dimensional space. Inspired by applications of attention mechanism for image processing, it can be inferred that, similar to image processing, information of critical local areas and feature channel of the point cloud has more impact on specific tasks.   

Based on these issues above, this paper proposes a point cloud feature extraction network, namely PointSCNet, which captures global structure and local region correlations of the point cloud for shape classification and part segmentation tasks. As shown in Figure~\ref{fig:point-order-attention}, a space filling curve guided sampling module is proposed to choose key points which represent geometric significant of local regions from the point cloud. Then, an information fusion module is designed to learn the structure and correlation information between those local regions. Moreover, a channel and spatial attention module is adopted for the final point cloud feature refinement. 

The main contributions of this paper are summarized as following:

 \begin{itemize} 
 \item An end-to-end point cloud processing network, namely PointSCNet, is proposed to learn structure and correlation information between local regions of a point cloud for shape classification and part segmentation tasks.
 \item The idea of space filling curve is adopted to points sampling and local sub-cloud generation. Specifically, points are encoded and sorted by Z-order curve coding, which makes the points contain meaningful geometric ordering.
 \item  An information fusion module is designed to represent local region correlation and shape structure information. The information fusion is achieved by correlating the local and structure feature via a correlation tensor, and skip connection operations.
 \item A channel-spatial attention module is adopted to learn the significant points and crucial feature channels. The channel-spatial attention weights are learned for the refinement of the point cloud feature.
 \end{itemize}

\section{Related Work}
\label{related_works}

This paper uses deep learning method to extract point cloud feature with construction and correlation information. In this section, recent researches in highly related area of our work, including traditional point cloud processing, point-wise embedding, point cloud structure reasoning and attention in point cloud processing, are briefly summarized and analyzed.

\subsection{Traditional point cloud processing methods}

One of the biggest challenges in processing point clouds is to deal with unstructured point cloud data. The early methods of processing point clouds are mostly indirect representation conversion. Some methods try to convert the point cloud to structured data, such as octree, kd-tree~\cite{klokov2017escape} to reduce the difficulty of analysis. Another classical kind of methods convert the point cloud to voxel models. The voxel-based methods~\cite{dai2017scannet,qi2016volumetric,wu20153d,johnson2017inferring} use 3D convolution, which is a direct extension of image processing applications for point cloud. The advantages of the methods are that they can preserve the spatial relationship well at high voxel resolution, but these methods are computationally very expensive. And if the resolution of voxelization is reduced, the geometric information that the voxels can represent will be significantly lost. FPNN~\cite{li2016fpnn}and Vote3~\cite{wang2015voting} proposed special methods to deal with the sparse problem, but their methods still cannot handle large-scale point cloud data well. Therefore, it is quite difficult to achieve real-time performance while considering the balance between accuracy and computational cost. the traditional methods inevitably lead to the loss of geometric information. This paper uses a point-by-point feature extraction method to overcome the high cost of voxel-based methods and is not conducive to processing low-resolution point clouds. 

\subsection{Point-wise embedding}

The research of PointNet~\cite{qi2017pointnet} is the breakthrough work for deep learning based directly point clouds processing method. Its groundbreaking proposal of max-pooling symmetric function solves the problem of disordered point clouds. The MLP layer extracts the features and uses the maximum pooling aggregation to obtain the global features of the point clouds. Then, the PointNet++~\cite{qi2017pointnet++} proposed a multi-layer sampling\&grouping method to improve the PointNet. A lot of later researches on point cloud processing~\cite{sun2019srinet,joseph2019momen,achlioptas2018learning, lin2019justlookup,zhang2020pointwise} followed the idea of point-wise and hierarchical point feature extraction. However, the feature extraction in the PointNet ignores geometrical structure information and the potential relationship between the local regions.
Therefore, in this paper, the point are embedded based on the idea of PointNet++ at first, and a space filling curve guided downsampling method and a information fusion method are proposed to learn the structure and correlation information of the point cloud.

\subsection{Point cloud structure reasoning}

 As an extension of point-wise feature leanring, various methods have been proposed to reason the structure of points. The DGCNN~\cite{wang2019dynamic} captures the features between point neighborhoods through graph convolution. The network extracts point cloud structure information by capturing the topological relationship between points. MortonNet~\cite{thabet2019mortonnet} proposed an unsupervised way to learn the local structure of point clouds. In PCT~\cite{guo2021pct}, the KNN method is adopted to extract the features between the point fields. The SRN ~\cite{duan2019structural} uses a concatenation for structural features and position coding between local sub-clouds, and the multi-scale features extracted by the method are used for point cloud processing, which improves the PointNet++~\cite{qi2017pointnet++}. However, these methods mainly pay attention to the relationship between local regions and ignore the relationship between the local region and the global shape. In this paper, a more effectively structure reasoning method is designed to capture the correlation between local regions and the shape structure.

\subsection{Attention in point cloud processing}
Due to the advancement of attention mechanism based method in many deep learning applications~\cite{lin2017structured,dosovitskiy2020image, woo2018cbam,vaswani2017attention}, the attention mechanism meets the demand of dealing with unstructured data and is well applied in point cloud processing~\cite{li2020unsupervised, guo2021pct, zhao2021point}. The Point Transformer~\cite{zhao2021point} and Point Cloud Transformer~\cite{guo2021pct} have made precedents for the application of the Transformer~\cite{vaswani2017attention} in point cloud processing and achieved the state-of-the-art performance. The adoption of attention mechanism for point cloud is mainly for exploring relation of points and enhancing the feature representation of attended points. Therefore, inspired by this idea, a channel-spacial attention module ~\cite{shaw2018self} is designed for feature refinement by enhancing key points and crucial feature channels.

\section{Method}
\label{methodoloy}

\begin{figure}[thb]
	\begin{center}
		\includegraphics[width=1\linewidth]{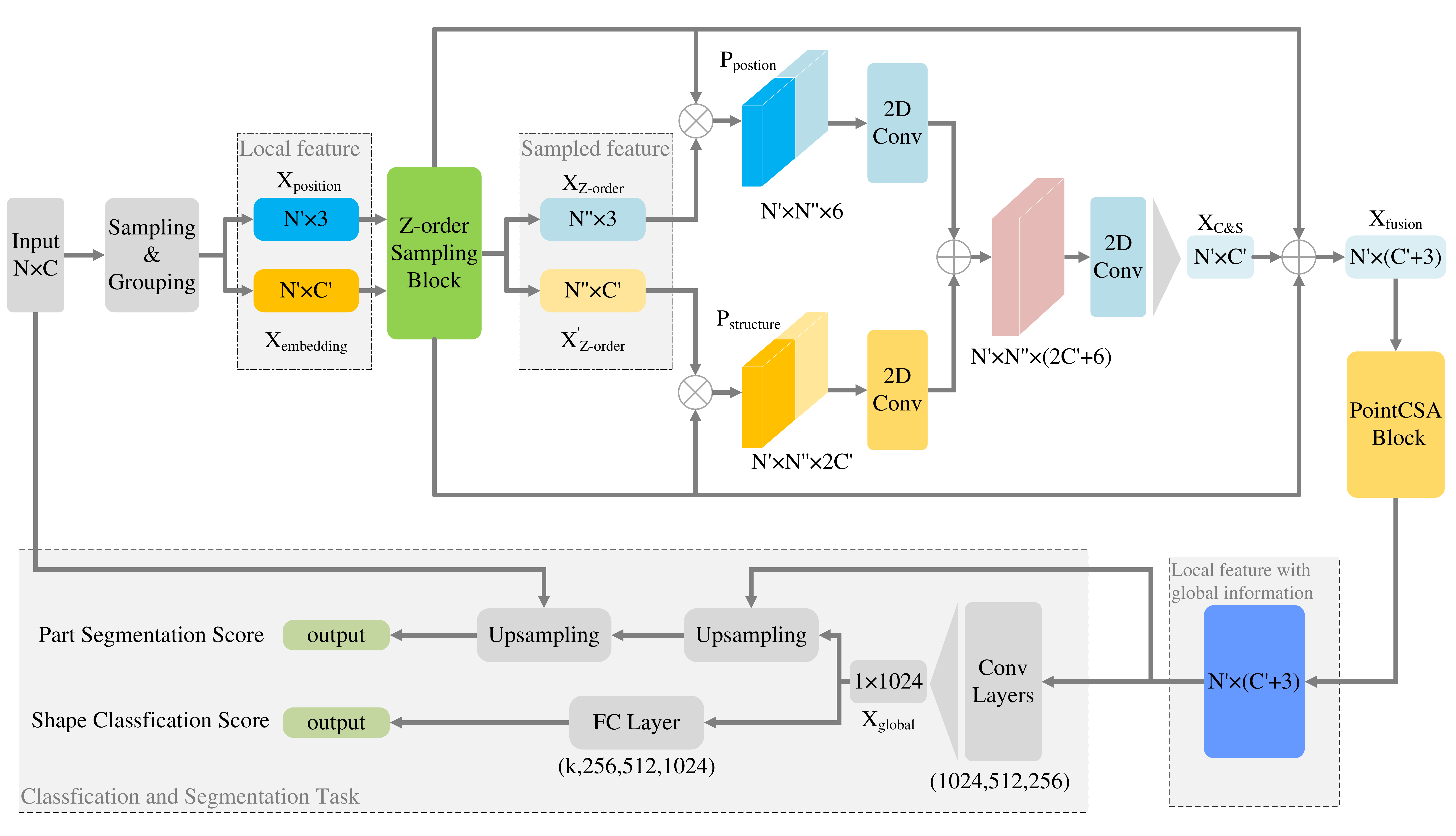}
	\end{center}
	\caption{Model architecture of PointSCNet: The original point cloud is fed to a sampling\&grouping block. Then a Z-order sampling block is designed for further generation of local regions. After the sampled point cloud feature is extracted, the feature fusion module is designed to learn the structure and correlation information. At last the point cloud feature is forwarded to the PointCSA block which is based on channel-spatial attention mechanism to get the refined feature for classification and segmentation.}
	\label{fig:Network}
\end{figure}

As shown in Figure~\ref{fig:Network}, the proposed PointSCNet first uses the original point set $P\in \mathbb{R}^{N\times C}$ as the input. $C$ is the feature channel of the point set. After a regular Sampling\&Grouping~\cite{qi2017pointnet++} block, we get the sampled point set of $N^{'}$ points with the original spatial position information, denoted as $X_{position}{\in}\mathbb{R}^{{N^{'}}{\times}3}$, and the embedded sampled point set with $C^{'}$ dimension feature, in which each point represents information of the surrounding points within a certain radius, denoted as $X_{embedding}{\in}\mathbb{R}^{{N^{'}}{\times}C^{'}}$.
Then we send $X_{embedding}$ and $X_{position}$  to a Z-order sampling module respectively for further sampling based on points' geometrical relation. The sampled point set contains the shape structure and local regions correlation information, denoted as $X_{Z-order}^{'}{\in}\mathbb{R}^{{N^{''}}{\times}C^{'}}$ and $X_{Z-order}{\in}\mathbb{R}^{{N^{''}}\times 3}$. 
After that, an information fusion module is designed to establish the correlation between each local sub-cloud and the entire point cloud for the shape structure and local region correlation information learning. 
Moreover, after the information fusion procedure, the point cloud feature is forwarded to a channel-spatial attention module for feature refinement.

The pipeline of classification and segmentation module is similar to the PointNet++~\cite{qi2017pointnet++}. The dimension of local point cloud features is increased to 1024 first, and then an aggregate function pooling is adopted to obtain $X_{global}{\in}\mathbb{R}^{1\times 1024}$ global features. For the shape classification task, after fed into the fully connected layers, the dimension of the global feature is reduced to $ 1\times k $ as the output of the PointSCNet, where $ k $ is the number of classes. For the part segmentation task, the output is the segmentation result $ N\times k^{'} $ obtained by up-sampling the global feature $X_{global}$, where $ k^{'} $ is the number of part classes.

\subsection{Initial Sampling{\rm{\&}}Grouping}
The PointSCNet first uses the original point cloud data as input. A series of points $X_{position}{\in}\mathbb{R}^{N^{'}\times 3}$ in the space are sampled via FPS, and the ball query method is used to get all points which are within a radius to the sampled point, denoted as

\begin{equation}
	\label{equ_fps}
	d(X_r, X_{position}) < r ,X_r{\in}\mathbb{R}^{N_{r}\times 3},
\end{equation}
where $X_{position}{\in}\mathbb{R}^{N^{'}\times 3}$ are the points sampled by FPS, $X_r{\in}\mathbb{R}^{N_{r}\times 3}$ are the points around $X_{position}$, $d( )$ is the Euclidean distance.

These points are encoded to a high-dimensional space through MLPs, and aggregated to the sampled point via the aggregation function $ Pooling() $ to get $ X_{embedding}\in \mathbb{R}^{N^{'}\times C^{'}} $,the aggregation function can be denoted as

\begin{equation}
	\label{equ_fps_pooling}
	X_{embedding}=Pooling(Concat(Conv(X_r), X_r))),
\end{equation}
where $X_{embedding}{\in}\mathbb{R}^{N^{'}\times C^{'}}$ are the encoded points feature, max-pooling function is used for pooling operation , the $Concat( )$ function represents points feature concatenation, the $Conv( )$ function is the 1D convolution operation.

After this procedure, the feature information of neighboring points has been aggregated to all sampled points $ X_{embedding} $.

\subsection{Z-order Curve Guided Sampling Module}

\begin{figure*}[h]
	\begin{center}
		\includegraphics[width=0.9\linewidth]{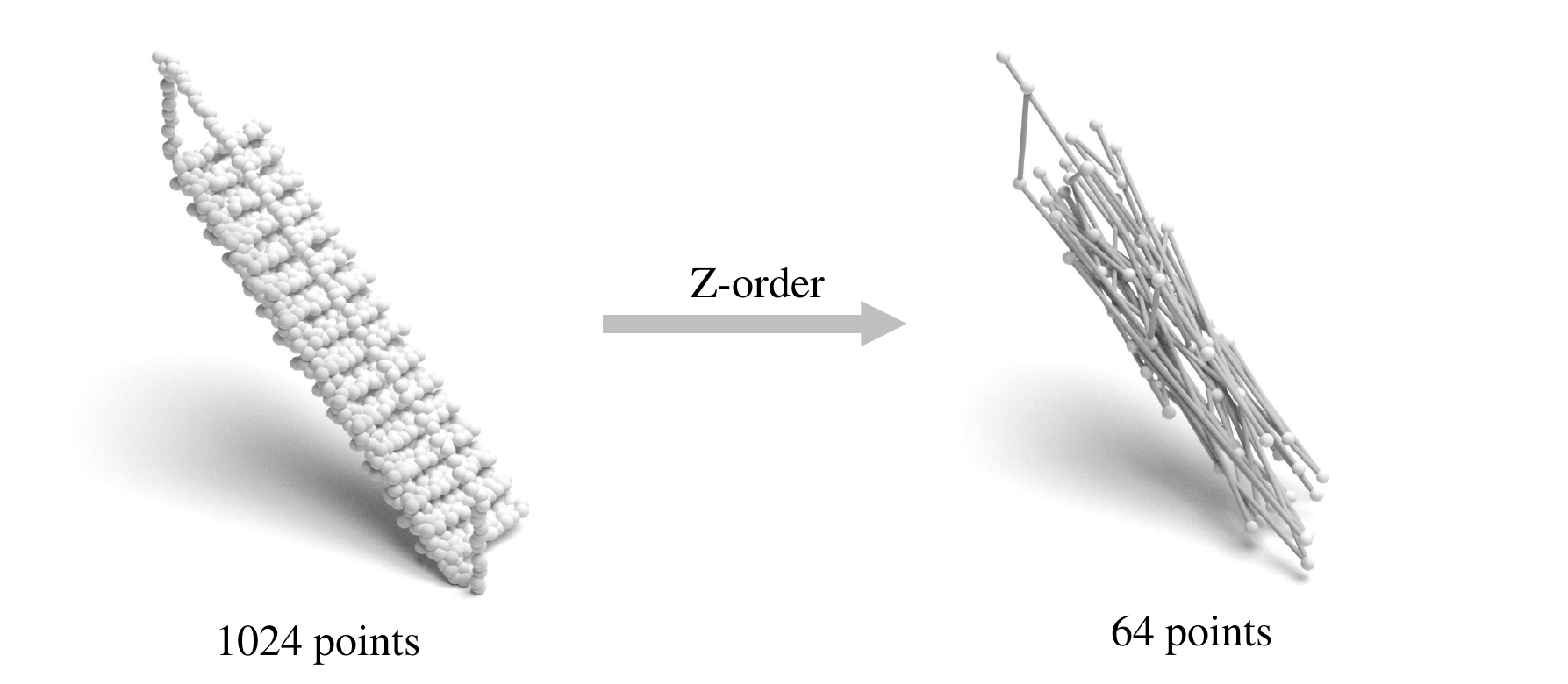}
	\end{center}
	\caption{The point cloud structure obtained by sampling 1024 points in the original point cloud using the Z-order space filling curve.}
	\label{fig:Zorderpic}
\end{figure*}

 The principle of space filling curve is to use a continuous curve to pass through all points in the space, and each point corresponds to a position code. After the FPS based sampling\&grouping, the Z-order curve coding function is adopted to further down-sampling the local sub-cloud $ X_{embedding} $ to get local regions with semantically high-level correlations.
 
\begin{figure*}[h]
	\begin{center}
		\includegraphics[width=0.8\linewidth]{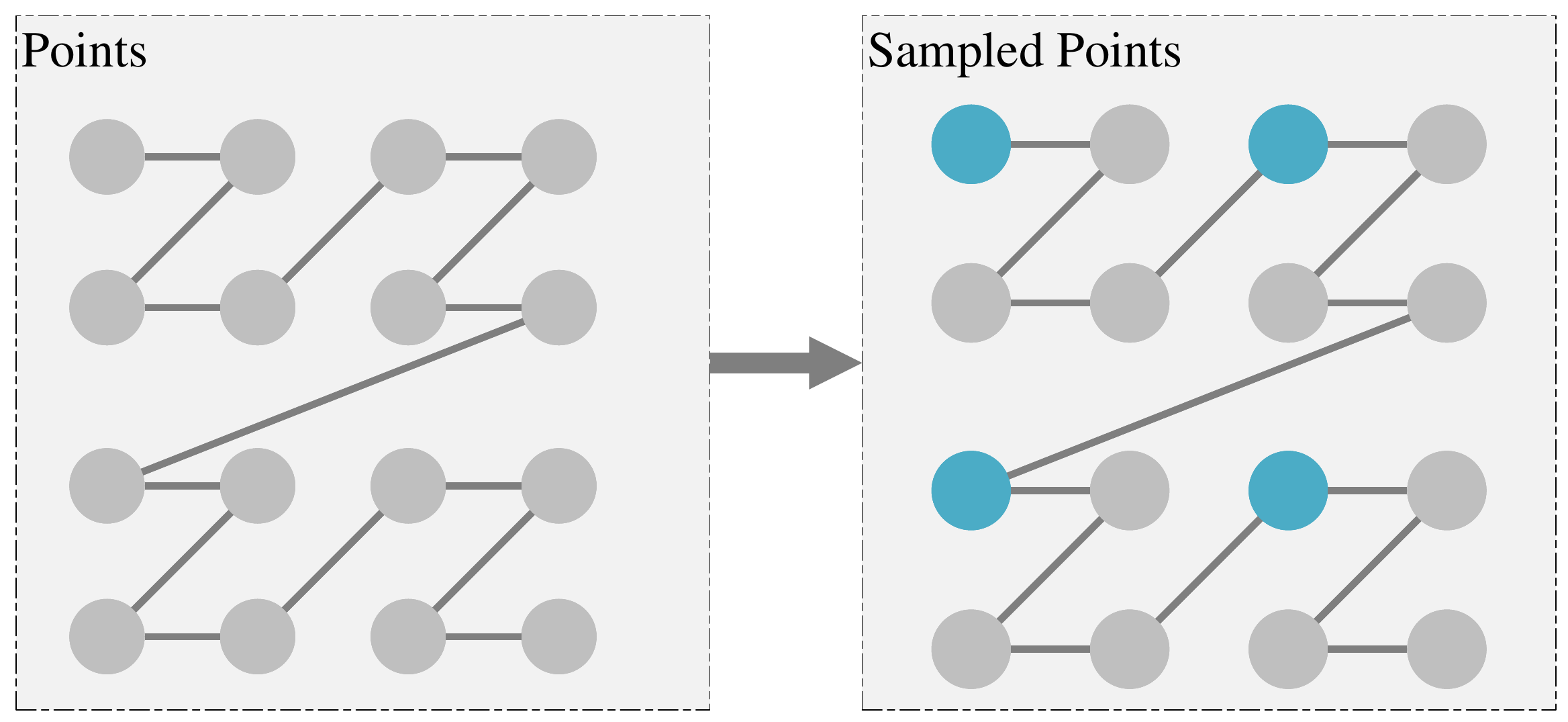}
	\end{center}
	\caption{Sampling strategy based on Z-order curve sorting. Equally spaced points are sampled, the spacing is set to 3 in the figure.}
	\label{fig:zorder-example}
\end{figure*}

After the Z-order encoding, the 3D position coordinates of the local sub-cloud will be mapped to the 1D feature space, as shown in Figure \ref{fig:Zorderpic}. 
The locality of the original point can be well-preserved due to the nature of Z-order curve, which means direct Euclidean neighbors in 1D tend to be similar to those in 3D. 
After the points are encoded and sorted, equally spaced points are sampled, as shown in Figure~\ref{fig:zorder-example}. 
Then, the point set with $N^{''}$ points and $C^{'}$ dimension feature, denoted as $X_{Z-order}^{'}{\in}\mathbb{R}^{{N^{''}}{\times}C^{'}}$, and the point set with $N^{''}$ points and 3D coordinate, denoted as $X_{Z-order}{\in}\mathbb{R}^{{N^{''}}\times 3}$ are sampled. The final sampled point set represents the global structure and local correlation of the original point set.

\subsection{Information fusion of local feature and structure feature}
\begin{figure*}[h]
	\begin{center}
		\includegraphics[width=0.85\linewidth]{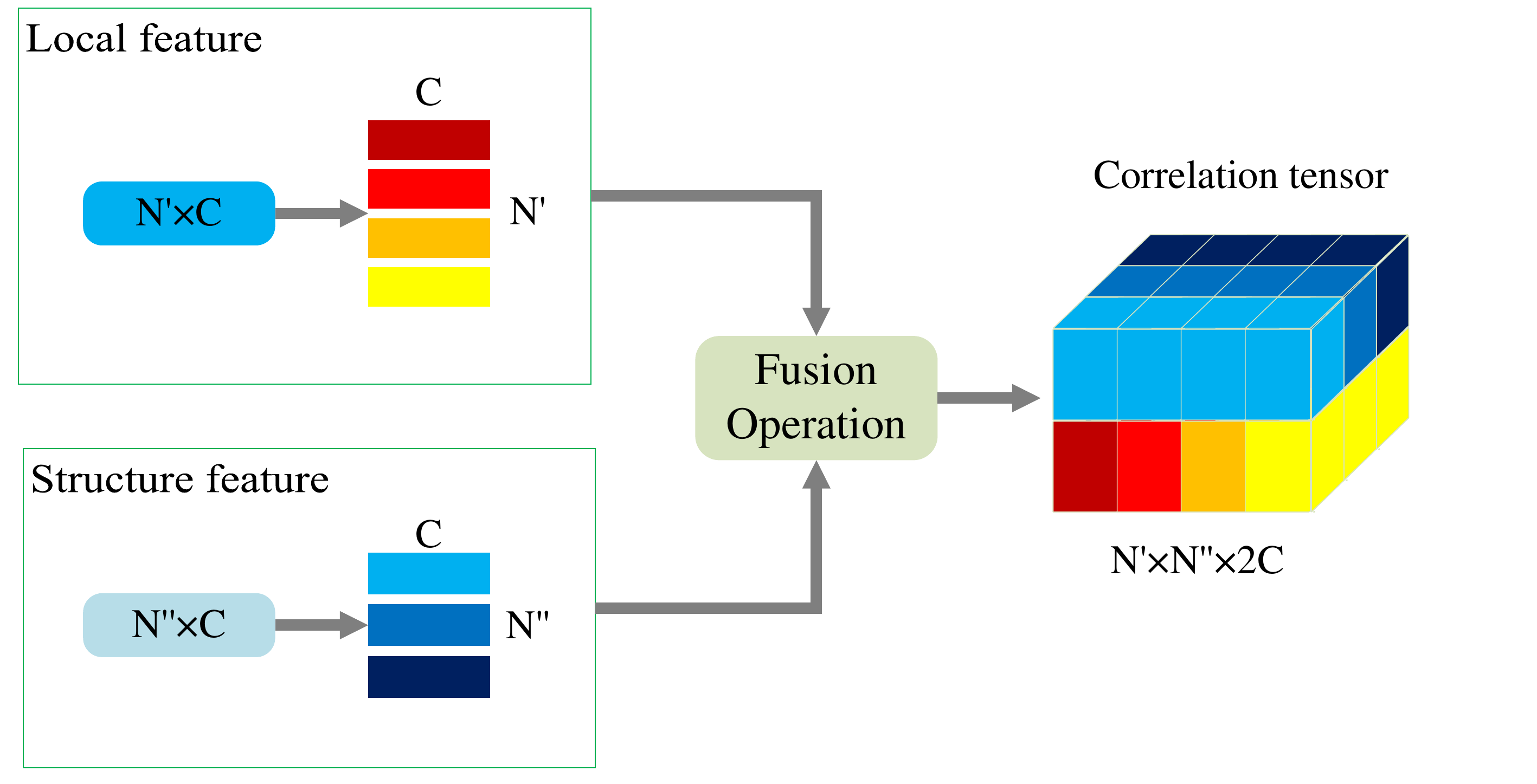}
	\end{center}
	\caption{The correlation tensor is designed for the evaluation of the correlation between the local feature and structure feature. $ N^{'} $ and $ N^{''} $ represent the number of points sampled via FPS and Z-order sampling block, $ C $ is the feature channel of points.}
	\label{fig:fusion-operation}
\end{figure*}

After obtained the Z-order based sampled point cloud, the local sub-cloud feature and the structure feature are correlated to learn the shape structure and local region correlation information. As shown in Figure~\ref{fig:fusion-operation}, a correlation tensor, represented as ${N^{'}\times{N^{''}}{\times}2C}$, is developed to evaluate the correlation between a local sub-cloud feature, represented as  $N^{'}{\times}C$ and a structure feature represented as  ${N^{''}{\times}C}$. The generation of the correlation tensor can be formalized as 

\begin{equation}
	\label{equ_concat_fusion}
	P_{structure} =Fusion(X_{embedding} , X_{Z-order}^{'}), P_{position} =Fusion(X_{position} , X_{Z-order}),
\end{equation}

\begin{equation}
	\label{matrix_fusion}
	Fusion(X,Y)=
	\left[
	\begin{matrix}
		Concat(X_1,Y_1)      & Concat(X_1,Y_2)     & \cdots & Concat(X_1,Y_n)      \\
		Concat(X_2,Y_1)      & Concat(X_2,Y_2)     & \cdots & Concat(X_2,Y_n)     \\
			\vdots				& \vdots			& \ddots & \vdots \\
		Concat(X_m,Y_1)      & Concat(X_m,Y_2)     & \cdots & Concat(X_m,Y_n)      \\
	\end{matrix}
	\right],
\end{equation}
where $X\in \mathbb{R}^{n\times c}$ and $Y\in \mathbb{R}^{m\times c}$, $ X,Y $ have the same numbers of feature channel. $X_i\in X$ and $Y_j\in Y$ are single point in point sets respectively. The $ Concat(  ) $ function is proposed to concatenate the feature channels of $X_i$ and $Y_j$.

Then 2D convolution layers are designed to obtain the structure and local correlation of the point cloud, as shown in Figure~\ref{fig:Network}. After the information fusion, a point cloud feature $X_{C\&S}{\in}\mathbb{R}^{{N^{'}}{\times}C^{'}}$ containing structure and correlation information is extracted. This process can be formalized as
\begin{equation}
	\label{equ_fusion}
	X_{C\&S} =H(P_{structure},P_{position})=Pooling(Relu(g(Concat(P_{structure},P_{position})))),
\end{equation} 
where $ g(  ) $ is the $ Conv2d $ function, $ Concat( ) $ is the concatenation operation.

Finally, as shown in Figure~\ref{fig:Network}, the$  X_{C\&S} $ , $ X_{embedding} $ and $ X_{position} $ are fused together to the fusion feature $ X_{fusion} $ via skip connections and the process can be formalized as 

\begin{equation}
	\label{equ_fusionconcat}
		X_{fusion}=Concat(X_{embedding}+X_{C\&S},X_{position}),
\end{equation}
where $X_{embedding}{\in}\mathbb{R}^{{N^{'}}{\times}C}$ represents local point cloud features, $X_{structure}{\in}\mathbb{R}^{{N^{'}}{\times}C}$ represents skeleton point cloud features, $X_{position}{\in}\mathbb{R}^{{N^{'}}{\times}3}$ represents  point cloud location features,the Concat function represents feature dimension concatenation of points.

\begin{figure}[t]
	\begin{center}
		\includegraphics[width=1\linewidth]{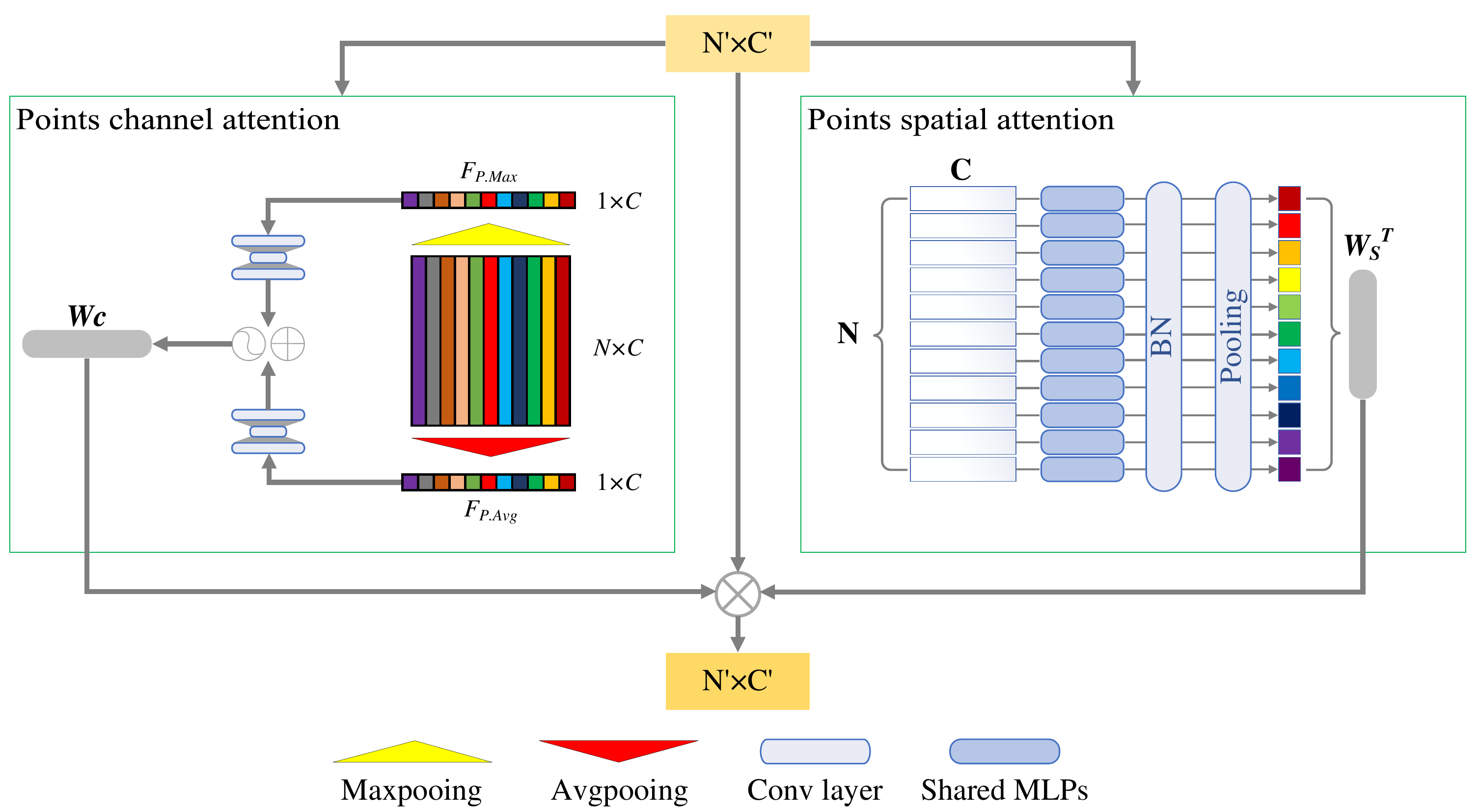}
	\end{center}
	\caption{Points channel-spatial attention module: The points feature is fed to the channel-spatial module to capture the most important points and feature channels. In channel attention module,channel weights are obtained via the two aggregation functions and convolution layers. In spatial attention module, spatial weights are obtained via shared MLPs.}
	\label{fig:CSAttention}
\end{figure}

\subsection{Points channel-spatial attention module}

As shown in Figure~\ref{fig:CSAttention}, a channel-spatial attention module which paralleling the channel attention block and the spatial attention block is adopted to strengthen the PointSCNet's ability by capturing the most important points and feature channels. In the channel attention module, the point cloud feature is aggregated by max-pooling and average pooling operation and then forwarded to convolution layers respectively. The design of the convolution layer is to reduce the feature dimension first and then raise it for better feature extraction. The outputs of convolution layers are summated and activated to learn the weight of each feature channel.      
The channel attention block is formalized as

\begin{equation}
	\label{equ_channel_attention}
	Channel(X)= ReLU(MLP(Max(X)+Avg(X))),
\end{equation}
where $X{\in}\mathbb{R}^{{N^{'}}{\times}C^{'}}$,$ Max( ) $ and $ Avg( ) $ represent max-pooling  and  average-pooling function.
In the spatial attention module, the feature is fed to the MLPs with shared weights, and then the information on each channel is aggregated through the batch normalization layer and the pooling layer to obtain the spatial position attention weight. The spatial attention block is formalized as

\begin{equation}
	\label{equ_spatial_attention}
	Spatial(X)= Pooling(BN(MLP(X))).
\end{equation}

\section{Experiments}
\label{experiments}

In this section, some quantitative and qualitative experiments are designed to demonstrate the performance of our proposed PointSCNet. At first, the network is evaluated on shape classification and part segmentation tasks respectively. Then, more quantitative analyses of the network are presented. Moreover, some more visualization experiments are performed to demonstrate the ability of PointSCNet quantitatively. Finally, the ablation study is designed to show the effectiveness of each module of PointSCNet.

\subsection{Implementation Details}

The development environment is Ubuntu18.04+Cuda11.1+Pytorch1.8.0, and the hardware environment is GPU device is $ RTX 3080 $ single discrete graphics. In the classification task, we set random sampling of $ 1024 $ points as the input of the PointSCNet, while random sampling of $ 2048 $ points in the segmentation task. Our training hyperparameters are set to the batch size of $ 24 $, the number of iterations is set to $ 200 $, the initial learning rate is set to $ 1e-3 $ , and the learning rate decays to the original $ 0.9 $ after every $ 20 $ iterations. The optimizer is Adam and weight decay rate is $1e-4$. In Z-order sampling, the number of sampled points is set to $ 64 $. The loss is measured by calculating the cross entropy between the real label and the predicted value. We set the number of local sub-clouds $ N^{'} =256 $, and the number of local sub-cloud feature channels $ C =192 $, The number of skeleton sub clouds $ N^{''}=64 $.

\subsection{Shape Classfication on ModelNet40}

The shape classification experiment is performed on the ModelNet40~\cite{wu20153d} dataset which is the most commonly used dataset for training point cloud classification networks. The dataset has 9,843 training data and 2,468 test data, belonging to 40 different shape classes. 
\begin{table}[h]
	\begin{center}
		\caption{Comparison with state-of-the-art methods on the ModelNet40 classification dataset. The column of "Acc" means overall accuracy(\%). All results quoted are taken from the cited papers. "xyz" in the column of Input means 3D coordinate of points and nr mean normals.}
		\begin{tabular}{p {2.3cm} p {1cm} p {1cm} p {1cm}}
			\toprule
			\makecell[l]{\textbf{Method}} & \makecell[c]{\textbf{Input}} &\makecell[c]{\textbf{points}} & \makecell[c]{\textbf{Acc}}\\
			\midrule
			\makecell[l]{Pointnet~\cite{qi2017pointnet}}&\makecell[c]{xyz}&\makecell[c]{1024}&\makecell[c]{89.2}\\
			\makecell[l]{Pointnet++~\cite{qi2017pointnet++}}&\makecell[c]{xyz}&\makecell[c]{1024}&\makecell[c]{90.7}\\
			\makecell[l]{Kd-Net~\cite{klokov2017escape}}&\makecell[c]{xyz}&\makecell[c]{32k}&\makecell[c]{91.8}\\
			\makecell[l]{DGCNN~\cite{phan2018dgcnn}}&\makecell[c]{xyz}&\makecell[c]{1024}&\makecell[c]{92.9}\\
			\makecell[l]{SRN~\cite{duan2019structural}}&\makecell[c]{xyz}&\makecell[c]{1024}&\makecell[c]{91.5}\\
			\makecell[l]{PointGrid~\cite{le2018pointgrid}}&\makecell[c]{xyz}&\makecell[c]{1024}&\makecell[c]{92.0}\\
			\makecell[l]{PointCNN~\cite{li2018pointcnn}}&\makecell[c]{xyz}&\makecell[c]{1024}&\makecell[c]{92.2}\\
			\makecell[l]{RS-CNN~\cite{liu2019relation}}&\makecell[c]{xyz}&\makecell[c]{1024}&\makecell[c]{93.6}\\
			\makecell[l]{PCT~\cite{guo2021pct}}&\makecell[c]{xyz}&\makecell[c]{1024}&\makecell[c]{93.6}\\
			\makecell[l]{PAConv~\cite{xu2021paconv}}&\makecell[c]{xyz}&\makecell[c]{1024}&\makecell[c]{93.9}\\
			\makecell[l]{CurveNet~\cite{muzahid2020curvenet}}&\makecell[c]{xyz}&\makecell[c]{1024}&\makecell[c]{93.8}\\
			\makecell[l]{RPNet-W9~\cite{ran2021learning}}&\makecell[c]{xyz}&\makecell[c]{1024}&\makecell[c]{93.9}\\
			\makecell[l]{Pointnet++~\cite{qi2017pointnet++}}&\makecell[c]{xyz,nr}&\makecell[c]{1024}&\makecell[c]{91.7}\\
			\makecell[l]{PAT~\cite{yang2019modeling}}&\makecell[c]{xyz,nr}&\makecell[c]{1024}&\makecell[c]{91.7}\\
			\makecell[l]{SpiderCNN~\cite{xu2018spidercnn}}&\makecell[c]{xyz,nr}&\makecell[c]{5k}&\makecell[c]{92.4}\\
			\makecell[l]{A-CNN~\cite{komarichev2019cnn}}&\makecell[c]{xyz,nr}&\makecell[c]{1024}&\makecell[c]{92.6}\\
			\makecell[l]{PointASNL~\cite{yan2020pointasnl}}&\makecell[c]{xyz,nr}&\makecell[c]{1024}&\makecell[c]{93.2}\\
			\makecell[l]{SO-Net~\cite{li2018so}}&\makecell[c]{xyz,nr}&\makecell[c]{1024}&\makecell[c]{93.4}\\
			\makecell[l]{\textbf{PointSCNet}}&\makecell[c]{xyz,nr}&\makecell[c]{1024}&\makecell[c]{\textbf{93.7}}\\
			\bottomrule
		\end{tabular}
		\label{com:Classfication table}
	\end{center}
\end{table}

In the shape classification experiment, we randomly sample 1024 point features as the input and 64 regions are sampled based on the Z-order curve coding. The PointSCNet is compared with some of the state-of-the-art methods. As shown in Table~\ref{com:Classfication table}, the overall classification accuracy of PointSCNet on ModelNet40 reaches 93.7\%, which outperforms or is on par with classical classification networks and recent state-of-the-art methods. 

PointNet++~\cite{qi2017pointnet++} is the pioneer work of hierarchical point cloud feature extraction, which captures the multi-scale local structure with hierarchical layers. It aggregates local features through a simple maximum pooling operation without using their structural relationship.  The DGCNN~\cite{phan2018dgcnn} and the SRN~\cite{duan2019structural} are both classical methods to learn structural relation of point cloud.  The DGCNN~\cite{phan2018dgcnn} simply concatenates the feature relationships of local sub-clouds in different dimensions, and the captured structural relationship features cannot fully represent the structure of the point cloud.The SRN~\cite{duan2019structural} adopts regular FPS based sampling\&grouping method to obtain local point cloud and simply concatenates the points position and geometry feature to capture the structure relationship.
The PointSCNet uses the space filling curve to sample the points in the point cloud that can characterize the point cloud structure, and then process them through a specially designed feature fusion module to explore the correlations between local regions and the structure of the point cloud. The performance is significantly improved compared to these baseline methods.

\begin{figure}[H]
	\begin{center}
		\includegraphics[width=1\linewidth]{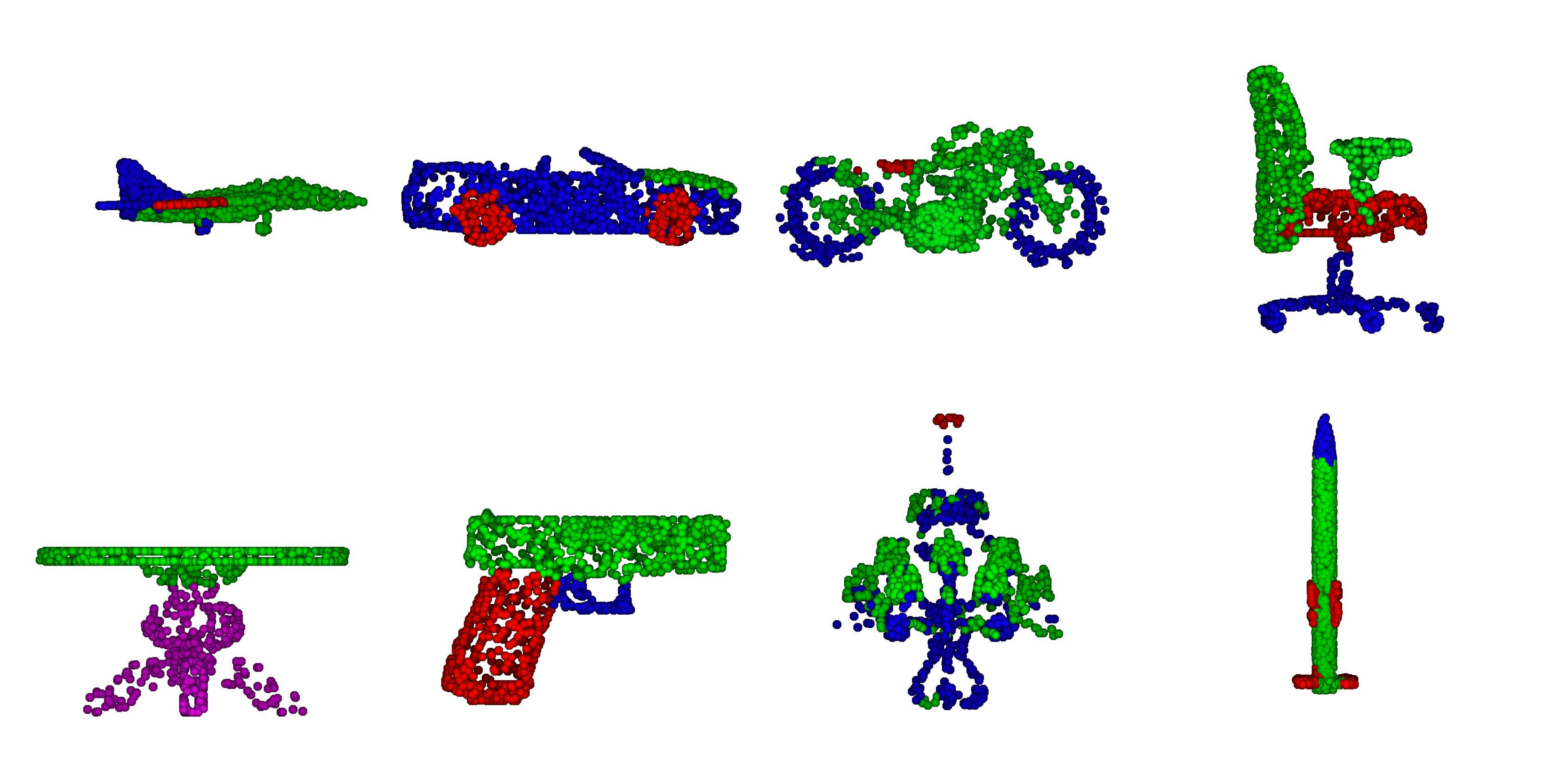}
	\end{center}
	\caption{Results of our PointSCNet on the part segmentation.}
	\label{fig:ShapeNet}
\end{figure}

\subsection{Part Segmentation on ShapeNet}

The ShapeNet~\cite{chang2015shapenet} dataset covers 55 common object categories and there are approximately 51,300 3D models. The part segmentation task is performed on the ShapeNet part segmentation dataset with 16,880 models and 16 categories. The 3D models are divided into 14,006 training point clouds and 2,874 test point clouds, where each point is associated with a point-by-point label of the point cloud segmentation task. In the point cloud component segmentation task, we randomly sampled 2048 point features as the original input of PointSCNet.  
The quantitative results of PointSCNet and some classical state-of-the-art methods are shown in table~\ref{com:Segficiation Compared}. By capturing the skeleton structure features of the point cloud, the PointSCNet significantly outperforms Pointnet~\cite{qi2017pointnet}, Pointnet++~\cite{qi2017pointnet++}, SRN~\cite{duan2019structural}, and the performance of PointSCNet is particularly outstanding in some specific classes, as shown in the table.
Figure~\ref{fig:ShapeNet} shows the visualization part segmentation result of PointSCNet.

\begin{table}[H]
	\begin{center}
		\caption{The performance of part segmentation task on ShapeNet. The metric is part-average Intersection-over-Union(IoU, \%). All results quoted are taken from the cited papers.}
		\begin{tabular}{p {1.6cm} p {1.8cm} p {2.2cm} p {1.3cm}p {1.7cm}p {2.1cm}p {1.9cm}}
			\toprule
			\makecell[l]{Class} & \makecell[c]{Pointnet~\cite{qi2017pointnet}} &\makecell[c]{Pointnet++~\cite{qi2017pointnet++}}&\makecell[c]{SRN~\cite{duan2019structural}}&\makecell[c]{PCNN~\cite{atzmon2018point}}&\makecell[c]{PointCNN~\cite{li2018pointcnn}}&\makecell[c]{\textbf{PointSCNet}}\\
			\midrule
			\makecell[l]{Airplane}&\makecell[c]{83.4}&\makecell[c]{82.3}&\makecell[c]{82.4}&\makecell[c]{82.4}&\makecell[c]{84.1}&\makecell[c]{83.3}\\
			\makecell[l]{Bag}&\makecell[c]{78.7}&\makecell[c]{79.7}&\makecell[c]{79.8}&\makecell[c]{80.1}&\makecell[c]{86.4}&\makecell[c]{84.3}\\
			\makecell[l]{Cap}&\makecell[c]{82.5}&\makecell[c]{86.1}&\makecell[c]{88.1}&\makecell[c]{85.5}&\makecell[c]{86.0}&\makecell[c]{\textbf{88.1}}\\
			\makecell[l]{Car}&\makecell[c]{74.9}&\makecell[c]{78.2}&\makecell[c]{77.9}&\makecell[c]{79.5}&\makecell[c]{80.8}&\makecell[c]{79.2}\\
			\makecell[l]{Chair}&\makecell[c]{89.6}&\makecell[c]{90.5}&\makecell[c]{90.7}&\makecell[c]{90.8}&\makecell[c]{90.6}&\makecell[c]{\textbf{91.0}}\\
			\makecell[l]{Earphone}&\makecell[c]{73.0}&\makecell[c]{73.7}&\makecell[c]{69.6}&\makecell[c]{73.2}&\makecell[c]{79.7}&\makecell[c]{74.3}\\
			\makecell[l]{Guitar}&\makecell[c]{91.5}&\makecell[c]{91.5}&\makecell[c]{90.9}&\makecell[c]{91.3}&\makecell[c]{92.3}&\makecell[c]{91.2}\\
			\makecell[l]{Knife}&\makecell[c]{85.9}&\makecell[c]{86.2}&\makecell[c]{86.3}&\makecell[c]{86.0}&\makecell[c]{88.4}&\makecell[c]{87.4}\\
			\makecell[l]{Lamp}&\makecell[c]{80.8}&\makecell[c]{83.6}&\makecell[c]{84.0}&\makecell[c]{85.0}&\makecell[c]{85.3}&\makecell[c]{84.5}\\
			\makecell[l]{Laptop}&\makecell[c]{95.3}&\makecell[c]{95.2}&\makecell[c]{95.4}&\makecell[c]{95.7}&\makecell[c]{96.1}&\makecell[c]{95.7}\\
			\makecell[l]{Motorbike}&\makecell[c]{65.2}&\makecell[c]{71.0}&\makecell[c]{72.2}&\makecell[c]{73.2}&\makecell[c]{77.2}&\makecell[c]{73.4}\\
			\makecell[l]{Mug}&\makecell[c]{93.0}&\makecell[c]{94.5}&\makecell[c]{94.9}&\makecell[c]{94.8}&\makecell[c]{95.3}&\makecell[c]{\textbf{95.3}}\\
			\makecell[l]{Pistol}&\makecell[c]{91.2}&\makecell[c]{80.8}&\makecell[c]{81.3}&\makecell[c]{83.3}&\makecell[c]{84.2}&\makecell[c]{81.7}\\
			\makecell[l]{Rocket}&\makecell[c]{57.9}&\makecell[c]{57.7}&\makecell[c]{62.1}&\makecell[c]{51.0}&\makecell[c]{64.2}&\makecell[c]{60.7}\\
			\makecell[l]{Skateboard}&\makecell[c]{72.8}&\makecell[c]{74.8}&\makecell[c]{75.9}&\makecell[c]{75.0}&\makecell[c]{80.0}&\makecell[c]{75.9}\\
			\makecell[l]{Mean}&\makecell[c]{83.7}&\makecell[c]{85.1}&\makecell[c]{85.3}&\makecell[c]{85.1}&\makecell[c]{86.1}&\makecell[c]{85.6}\\
			\bottomrule
		\end{tabular}
		\label{com:Segficiation Compared}
	\end{center}
\end{table}

\subsection{Additional Quantitative Analyses}

\begin{figure}[h]
	\begin{center}
		\includegraphics[width=0.5\linewidth]{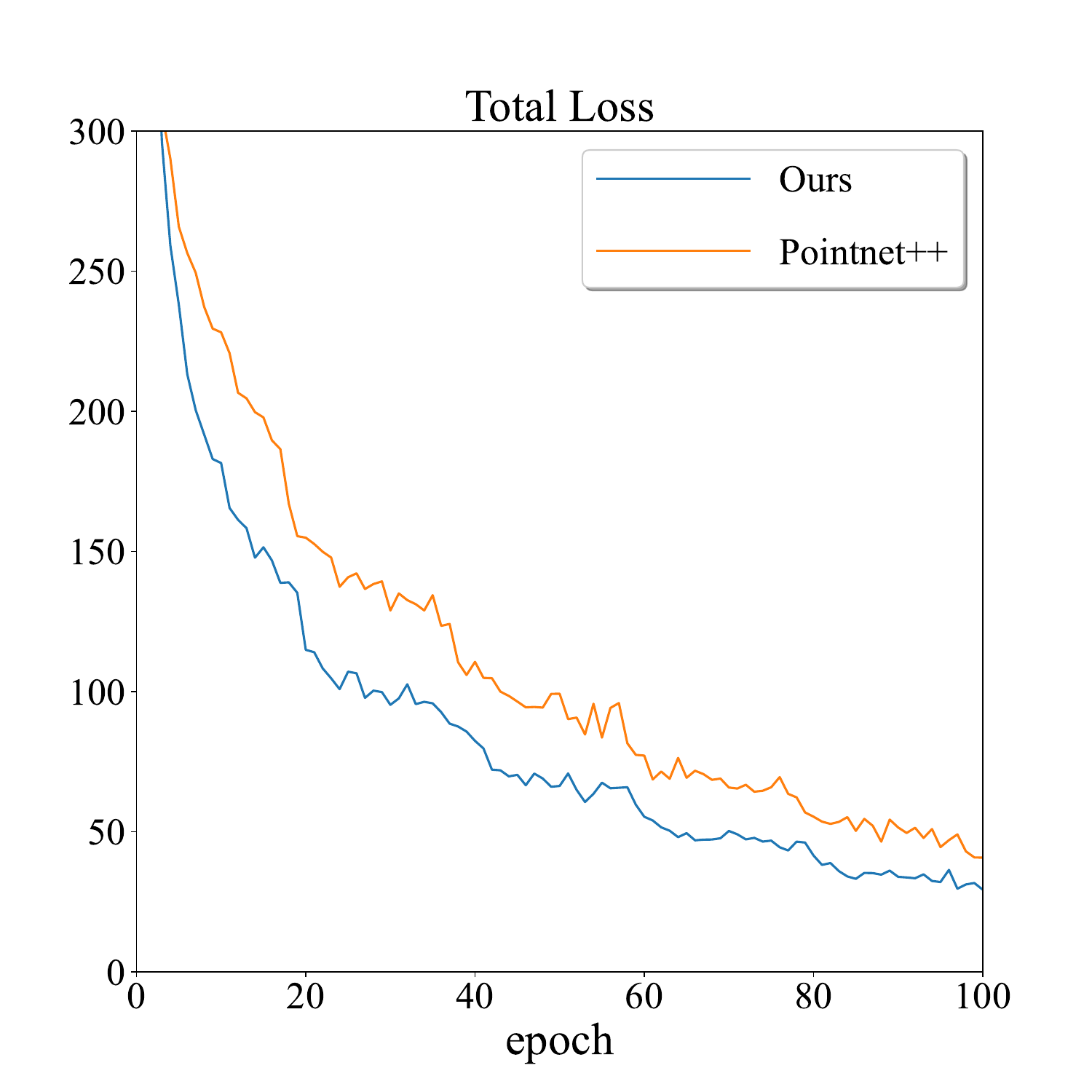}
	\end{center}
	\caption{Experiment on the drop speed of loss curve.}
	\label{fig:Dropspeed}
\end{figure}

The number of model parameters reflect the training speed of the network indirectly. 
Our PointSCNet adopts the space filling curve guided sampling strategy to capture few points to represent local regions and structure of the point cloud, which reduces the number of model parameters. The PointSCNet achieves outstanding classification accuracy with relatively few model parameters, as shown in table~\ref{com:Parameters Compared}.
For the PointNet~\cite{qi2017pointnet++}, its multi-layer sampling structure introduces redundant information and slow down the training speed. Figure~\ref{fig:Dropspeed} shows the loss curve of PointSCNet decreases more rapidly compared to the Pointnet++.
The PCT~\cite{guo2021pct} network uses the Transformer structure repeatedly to capture the structural relationship characteristics of the point cloud. Hence it has excessive parameters and its convergence speed is slow.
The SRN~\cite{duan2019structural} adopts regular FPS based sampling\&grouping method to obtain sub regions and adopts duplicated SRN module, which leads to a large number of parameters and a slow convergence speed too.

\begin{table}[H]
	\begin{center}
		\caption{The performance of PointSCNet on the ModelNet40 dataset to test classification tasks}
		\begin{tabular}{p {2.3cm} p {1.5cm} p {1cm}}
			\toprule
			\makecell[l]{\textbf{Method}} & \makecell[c]{\textbf{Params}} &\makecell[c]{\textbf{Acc}}\\
			\midrule
			\makecell[l]{Pointnet~\cite{qi2017pointnet}}&\makecell[c]{3.472M}&\makecell[c]{89.2}\\
			\makecell[l]{Pointnet++~\cite{qi2017pointnet++}}&\makecell[c]{1.748M}&\makecell[c]{91.9}\\
			\makecell[l]{SRN~\cite{duan2019structural}}&\makecell[c]{3.743M}&\makecell[c]{91.5}\\
			\makecell[l]{DGCNN~\cite{phan2018dgcnn}}&\makecell[c]{1.811M}&\makecell[c]{92.9}\\
			\makecell[l]{NPCT~\cite{guo2021pct}}&\makecell[c]{1.36M}&\makecell[c]{91.0}\\
			\makecell[l]{SPCT~\cite{guo2021pct}}&\makecell[c]{1.36M}&\makecell[c]{92.0}\\
			\makecell[l]{PCT~\cite{guo2021pct}}&\makecell[c]{2.88M}&\makecell[c]{93.2}\\			
			\makecell[l]{\textbf{PointSCNet}}&\makecell[c]{\textbf{1.827M}}&\makecell[c]{\textbf{93.7}}\\
			\bottomrule
		\end{tabular}
		\label{com:Parameters Compared}
	\end{center}
\end{table}

\subsection{Additional Visualization Experiments}
\begin{figure}[h]
	\begin{center}
		\includegraphics[width=1\linewidth]{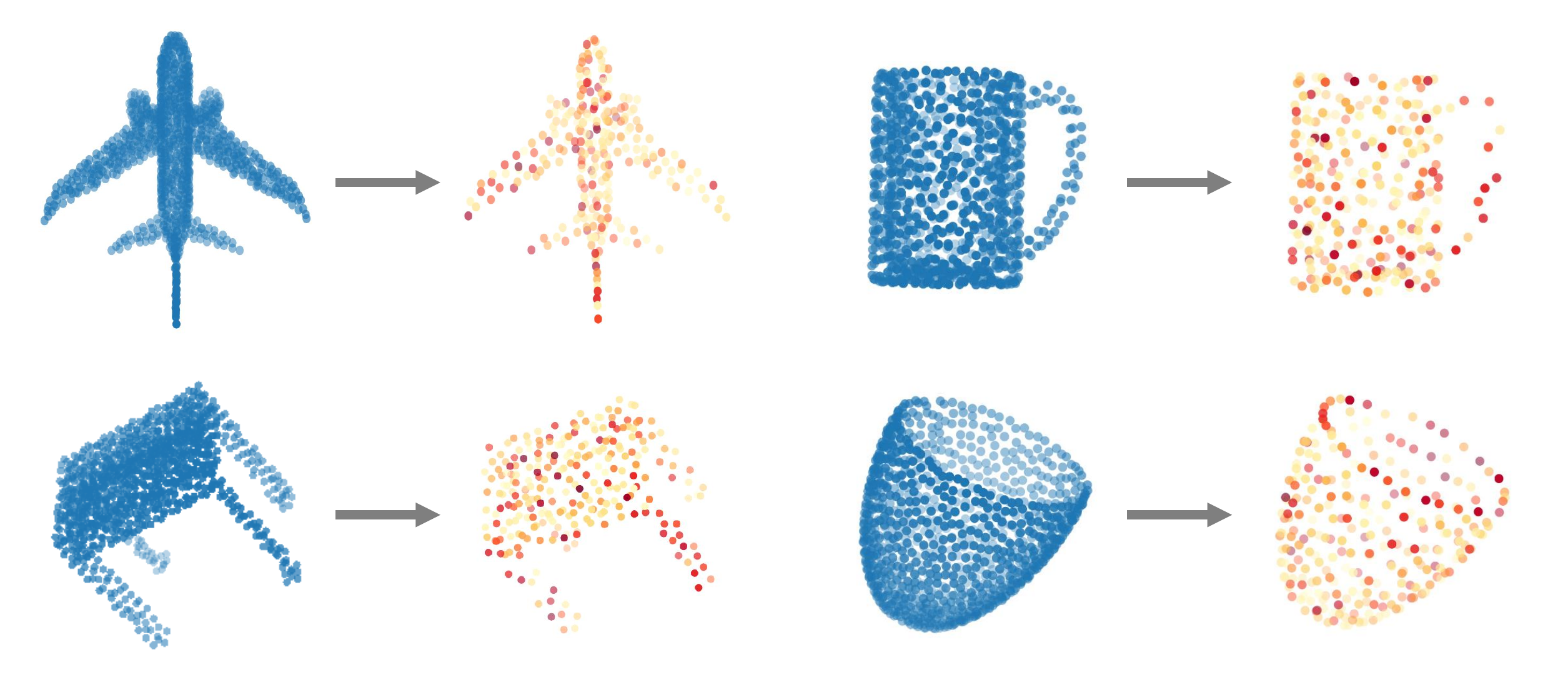}
	\end{center}
	\caption{Heat map for points with high response to PointSCNet.}
	\label{fig:symmetry}
\end{figure}

The heat map for points with high response to PointSCNet is shown in Figure~\ref{fig:symmetry}. The points are colored according to their response to the network and those with higher response are colored darker. The darker points in of the mug display the model structure. The darker points in the airplane mainly gather in one side of the symmetry axis, which indicates the symmetry of the airplane model. In the table model, both model structure and repetitive arrayed table leg are emphasized. The points with high response appear on the circle rim of the bowl. 

According to the visualization results, those points with higher response either present the structure of a point cloud or show the geometrical and locational interactions of local regions, which proves the points sampled by the Z-order sampling module represents the meaningful geometrical local regions and the information fusion block extracts the structure and correlation information effectively.        



\begin{figure}[H]
	\begin{minipage}[t]{0.6\textwidth}
		\centering	
		\includegraphics[width=10cm]{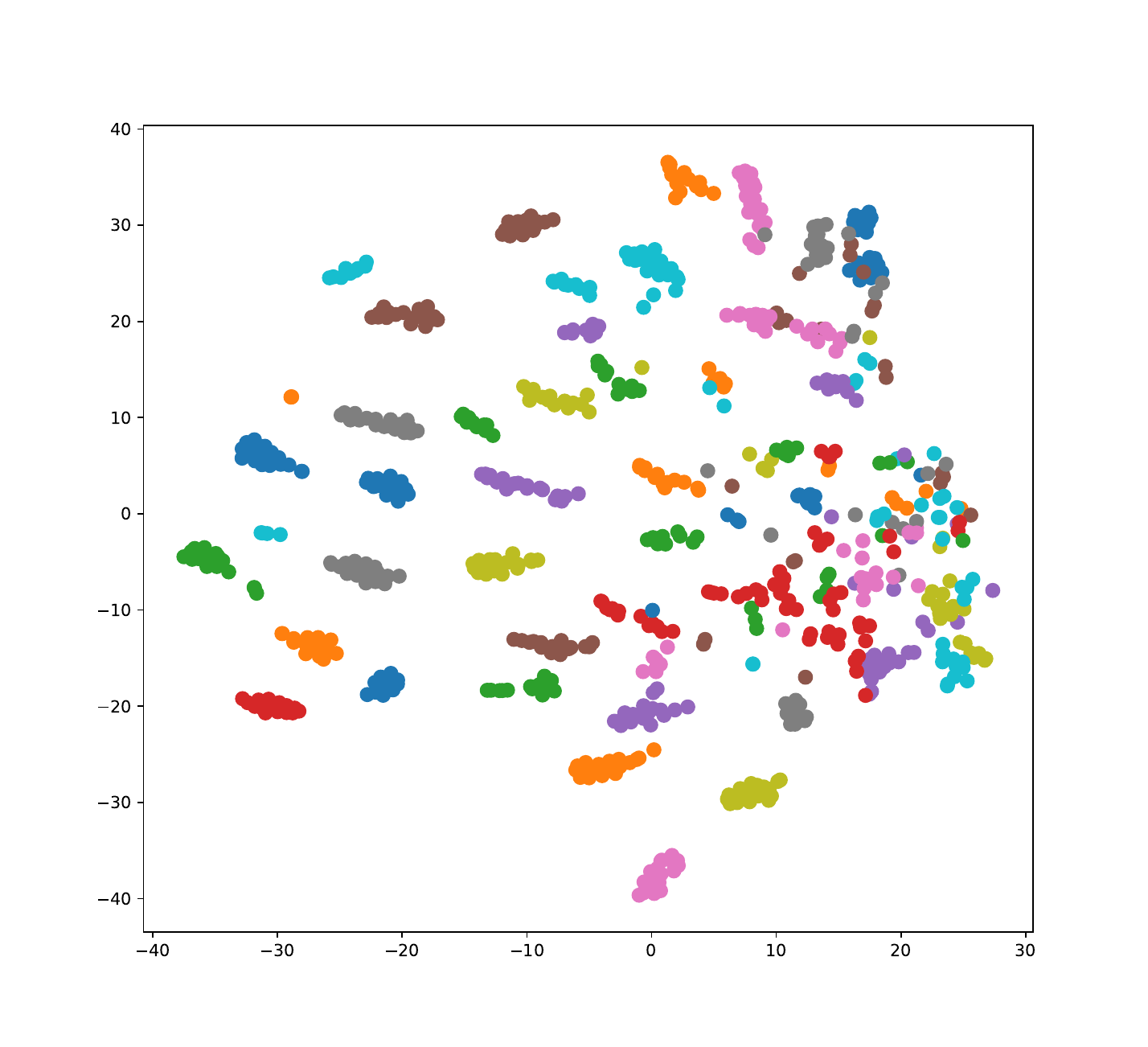}	
	\end{minipage}
	\begin{minipage}[t]{0.3\textwidth}
		\flushleft
		\raisebox{0.08\height}{\includegraphics[width=4.5cm]{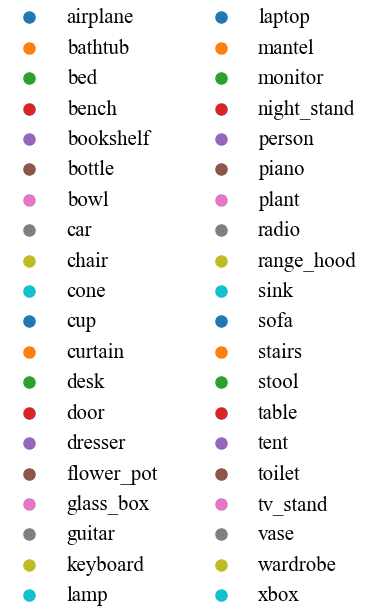}}
	\end{minipage}
	\caption{Visualization results of t-SNE on the ModelNet40 dataset.}
	\label{fig:TSNE}
\end{figure}

Figure~\ref{fig:TSNE} shows the performance of our PointSCNet in feature extraction. By using t-SNE~\cite{van2008visualizing} to reduce the dimension of high-dimensional features to 2D, the classification ability of our network is visualized as shown in the figure. It can be seen that most of the classes are divided into separate clusters. For some clusters with similar point cloud structures, such as tables and stools are close in semantic space, the PointSCNet can still distinguish them precisely.

\subsection{Ablation Study}

A set of ablation studies are designed to test the impacts of critical components of our network, including the Z-order sampling block (section 3.2), structure and correlation information fusion module (section 3.3) and the channel-spatial attention block (section 3.4). The ablation strategies and results are shown in Table~\ref{com:Ablation Studies}. It can be found that all the critical components of the PointSCNet improve the network performance. The Z-order sampling block and C\&S module provide obvious improvement. The convergence speed is slow while only use the information fusion module. When all these three modules are used at the same time, the model training speed is greatly improved, and the highest accuracy of the classification task is achieved, which further proves the importance of each module.
\begin{table}[H]
	\begin{center}
		\caption{The strategies and results of ablation studies. "ZS" represents the Z-order curve guided sampling block. "C\&S" represents the structure and correlation information fusion module. "AM" is the channel-spatial attention module. "\checkmark" represents existence, "$\times$" represents inexistence. "ToBestAcc" is the minimum number of the epoch when the PointSCNet achieve the highest accuracy in the training phase.}
		\begin{tabular}{p {2cm} p {1cm} p {1cm} p {1cm}p {1cm}p {3cm}}
			\toprule
			\makecell[c]{Methods} & \makecell[c]{ZS} &\makecell[c]{C\&S}&\makecell[c]{AM}&\makecell[c]{Acc}&\makecell[c]{ToBestAcc/epochs}\\
			\midrule
			\makecell[c]{A}&\makecell[c]{\checkmark}&\makecell[c]{$\times$}&\makecell[c]{$\times$}&\makecell[c]{93.0}&\makecell[c]{87}\\
			\makecell[c]{B}&\makecell[c]{\checkmark}&\makecell[c]{\checkmark}&\makecell[c]{$\times$}&\makecell[c]{93.4}&\makecell[c]{95}\\
			\makecell[c]{C}&\makecell[c]{\checkmark}&\makecell[c]{$\times$}&\makecell[c]{\checkmark}&\makecell[c]{93.2}&\makecell[c]{85}\\
			\makecell[c]{D}&\makecell[c]{$\times$}&\makecell[c]{\checkmark}&\makecell[c]{\checkmark}&\makecell[c]{93.3}&\makecell[c]{120}\\
			\makecell[c]{E}&\makecell[c]{$\times$}&\makecell[c]{\checkmark}&\makecell[c]{$\times$}&\makecell[c]{93.2}&\makecell[c]{148}\\
			\makecell[c]{F}&\makecell[c]{$\times$}&\makecell[c]{$\times$}&\makecell[c]{$\checkmark$}&\makecell[c]{93.2}&\makecell[c]{73}\\
			\makecell[c]{PointSCNet}&\makecell[c]{\checkmark}&\makecell[c]{\checkmark}&\makecell[c]{\checkmark}&\makecell[c]{93.7}&\makecell[c]{67}\\
			\bottomrule
		\end{tabular}
		\label{com:Ablation Studies}
	\end{center}
\end{table}

\section{Conclusion}
\label{conclusion}

In this paper, a point cloud processing network named PointSCNet is proposed to learn the shape structure and local region correlation information based on space filling curve guided sampling. Different from most existing methods using FPS method for down-sampling, which only utilizes the low-dimension Euclidean distance, our proposed space filling curve guided sampling module uses the Z-order curve for sampling to explore high-level correlations of points and local regions. The feature of sampled points are fused in the proposed information fusion block, in which the shape structure and local region correlation are learned. Finally, the channel-spatial module is designed to enhance the feature of key points. Quantitative and qualitative experimental results demonstrate that the proposed PointSCNet learns the point cloud structure and correlation effectively, and achieves superior performance on shape classification and part segmentation tasks. The idea of structure and correlation learning can be adopted for related vision tasks other than 3D points processing. Hence, in future we plan to optimize our network and apply the method to more vision scenarios~\cite{zhang2020part,zhang2021deep,wu20203d}.

\vspace{6pt} 



\authorcontributions{X.C. and Y.W. conceived and designed the algorithm and the experiments. X.C. wrote the manuscript. Y.W. supervised the research. W.X. assisted in the experiments. Y.W. provided suggestions for the proposed method and its evaluation and assisted in the preparation of the manuscript. W.X. and Y.C assisted in the experiments. J.L. analyzed the data. H.D. collected and organized the literatures. All authors have read and agreed to the published version of the manuscript.}

\funding{This work is supported by the National Science Foundation of China (Grant No. 61802355 and 61702350) and Hubei Key Laboratory of Intelligent Robot (HBIR 202105).}

\conflictsofinterest{The authors declare no conflict of interest.} 

\reftitle{References}


\bibliography{Definitions/reference1}




\end{document}